# Twin Systems for DeepCBR: A Menagerie of Deep Learning and Case-Based Reasoning Pairings for Explanation and Data Augmentation

**Mark T. Keane,**[1,2,3] **Eoin M. Kenny,**[1,3] **Eoin Delaney,**[1,3] **Mohammed Temraz,**[1,3]
**Derek Greene**[1,2,3] and **Barry Smyth**[1,2]

[1]School of Computer Science, University College Dublin, Dublin, Ireland
[2]Insight Centre for Data Analytics, Dublin, Ireland
[3]VistaMilk SFI Research Centre, Ireland
{mark.keane, derek.greene, barry.smyth}@ucd.ie,
{mohammed.temraz, eoin.kenny1, eoin.delaney.4}@ucdconnect.ie

## Abstract

Recently, it has been proposed that fruitful synergies may exist between Deep Learning (DL) and Case-Based Reasoning (CBR), that there are insights to be gained by applying CBR ideas to problems in DL (what could be called DeepCBR). In this paper, we report on a program of research that applies CBR solutions to the problem of Explainable AI (XAI) in DL. We describe a series of twin-systems pairings of opaque DL models with transparent CBR models that allow the latter to explain the former using factual, counterfactual and semi-factual explanations. This twinning shows that functional abstractions of DL models (e.g., from feature weights, feature contributions, predictive analyses) can be used to build explanatory solutions. We also float the idea that some of these techniques may apply to the problem of Data Augmentation in DL, underscoring the fecundity of these DeepCBR ideas.

## 1 Introduction

Recently, Leake and Crandall [2020] have argued that key challenges facing Deep Learning (DL) could benefit from insights arising from Case-Based Reasoning (CBR). Adopting this perspective, we have mined a rich vein of research based on applying CBR to the challenges raised by eXplainable AI (XAI) for Deep Learning (see e.g., [Keane and Kenny, 2019; Keane and Smyth, 2020; Keane *et al.,* 2021; Kenny and Keane 2019, 2021a, 2021b; Kenny *et al.,* 2019, 2021; Delaney *et al.,* 2021; Smyth and Keane, 2021]), along with recent extensions to Data Augmentation [Temraz and Keane, 2021; Temraz *et al.,* 2021]. This work started from long-standing proposals in CBR on the use of factual, case-based explanations [Leake and McSherry, 2005; Sørmo *et al.,* 2005], but has been extended to consider the use of counterfactual and semi-factuals explantions as well.

This work is anchored by the notion of *Twin Systems*, in which an opaque black box (DL) model is mapped to a more transparent (CBR) model, to allow the latter to explain the former (see Figure 1 and [Keane and Kenny, 2019]). In this paper, we review our recent work on applying this idea to the XAI challenges in DL, we update the definition of twinning and consider the novel departure of applying twinning to data augmentation. As such, this paper profiles what we hope to be, a set of interesting and productive ideas for how CBR can be used to benefit, support and expand DL.

In the remainder of this introduction we (i) introduce three explanatory strategies for DL based on factual, counterfactual and semi-factual cases, (ii) update our earlier definition of twin systems [Kenny and Keane, 2019, 2021b; Kenny *et al.,* 2021], and (iii) outline the structure for the rest of the paper.

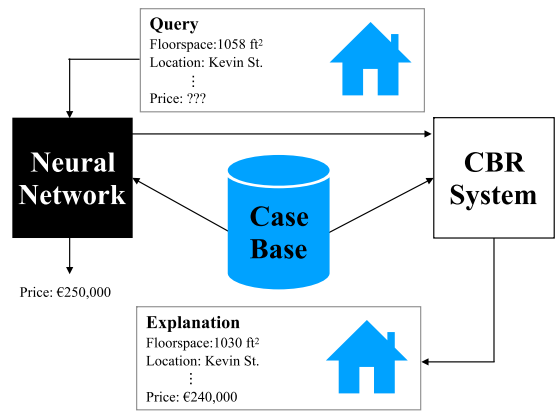

**Fig. 1**: A simple ANN-CBR twin-system (adapted from [Kenny and Keane, 2019]); a query-case posed to an ANN gives an accurate, but unexplained, prediction for a house price. The ANN is twinned with the CBR system (both use the same dataset), and its feature-weights are mapped to the CBR model to retrieve an example-based nearest-neighbor case to explain the ANN's prediction.

## 1.1 A Tale of Three Explanations

Recently, the XAI literature has rapidly moved from the more traditional factual, example/case-based explanations to counterfactual [Byrne, 2019; Miller, 2020; Karimi *et al.,* 2020; Keane *et al.,* 2021] and semi-factual explanations [Kenny and Keane, 2021a]. Taxonomically, these reflect options for post-hoc explanations using different types of examples. Figure 2 shows the main divisions in approaches to XAI along with some solutions (blue ovals), reflecting recent proposals on XAI taxonomies [see e.g., Lipton, 2018].

**Factual Explanations**. These explanations are the case-based examples discussed in hundreds, if not thousands of CBR papers [Leake and McSherry, 2005; Sørmo *et al.,* 2005]; except that now the example-cases to explain DL are retrieved guided by extracted feature-weighs from the DL [Kenny and Keane, 2019]. Imagine a SmartAg system, where a DL model for crop growth tells a farmer than in the next week, the grass yield on their farm will be 23 tons, and the farmer asks "Why?" [Kenny *et al.,* 2019]. Using these techniques, a factual explanation could be found from historical cases for this farm using its case base, replying "Well, next week is like week-12 from two years ago, in terms of the weather and your use of fertilizer and that week yielded 22.5 tons of grass". This explanatory factual case comes from finding the nearest neighbor in the CBR's dataset (aka the training data for the DL). Kenny *et al.* [2021] tested a twin-system of this type, in user studies using image-data and found that people's perceptions of incorrect items are improved by explanations, though the explanations do not mitigate people's negative assessment of the model, when it makes errors in its predictions.

**Counterfactual explanations**. This explanation type is quite different to the factual option. It tells the end-user about how things would have to change for the model's predictions to change (hence, it can be used for algorithmic recourse; Karimi *et al.* [2020]). Imagine the farmer thinks that the crop yield should be higher than 23 tons and asks, "Why not?"; now, the AI could provide advice for getting a better yield in the future, by explaining that "If you doubled your fertilizer use, then you could achieve a higher yield of 28 tons". So, unlike factual explanations which tend to merely justify the status quo, counterfactuals can provide a basis for actions that can change future outcomes (see [Byrne, 2019; Miller, 2019] on the psychology of counterfactual explanations for XAI).

**Semi-factual explanations**. Finally, semi-factual explanations also have the potential to guide future actions. Imagine again, the farmer thinks that the crop yield should be higher than 23 tons and asks, "Why not?"; now, the AI could provide a semi-factual "even-if" explanation that is also quite informative saying "*Even if* you doubled your fertilizer use, the yield would still be 23 tons". In this case, the farmer is potentially warned-off over-fertilizing and creating conditions that might pollute the environment. Semi-factuals have been examined occasionally in psychology [McCloy and Byrne, 2002], but hardly at all in AI (see Nugent *et al's*

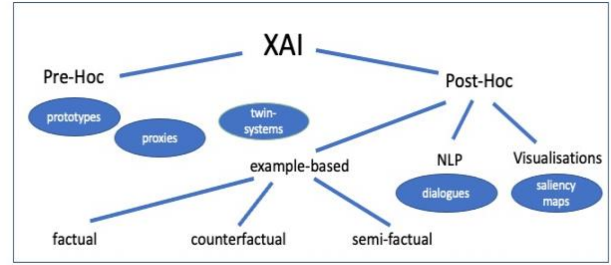

**Fig 2**. Broad taxonomic divisions in Explainable AI (XAI)

[2009] discussion of *a fortori reasoning* for one notable exception in the CBR literature).

As we shall, these three options for explanation have been explored within the twin-systems approach, in which CBR ideas are used to make DL more transparent.

## 1.2 Re-Defining Twinning and Twins

Keane and Kenny [2019] reviewed a potential synergy between DL and CBR models for XAI, coining the term "twin systems" to describe a framework for them (see Figure 1). After considering the literature, they found sporadic research on using CBR to explain Artificial Neural Networks [ANNs, then Multi-layered Perceptrons (MLPs)] from the 1990s to 2000s [Nugent *et al.,* 2009; Shin *et al.,* 2000]; this research explored the idea that case-based examples could provide good explanations [Leake and McSherry, 2005; Sørmo *et al.* 2005]. These systems paired an opaque ANN with a more transparent CBR system for explanatory purposes (i.e., an ANN-CBR twin). Typically, this twinning extracted feature weights from the ANN using various methods and used them in the CBR retrieval step to find factual, explanatory cases (see Figure 1). Recently, this general approach has been extended to other explanation strategies (e.g., counterfacutals) and DL architectures [e.g., convolutional neural networks (CNNs)]. Looking across these diverse efforts, the original definition of twin systems needs to be extended to reflect these developments. Accordingly, we re-define ANN-CBR twins as having (the underlined words show the changes to the wording of the original definition):

- *Two Techniques*. A hybrid system where an ANN (a MLP or DL model) and a CBR technique (notably, a *k*-NN) are combined to meet the system requirements of accuracy and interpretability.
- *Separate Modules*. These techniques are run as separate, independent modules, "side-by-side".
- *Common Dataset*. Both techniques are applied to the same dataset (i.e., twinned by this common usage).
- *Functionality Mapping*[1]. Some characterization of the ANN's functionality – typically described as *feature-*

---
[1] Previously, this part of the definition had the heading "Feature-Weight Mapping", which is now generalized somewhat.

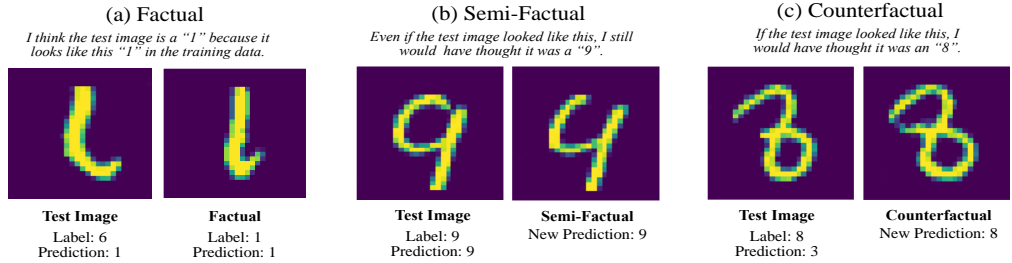

**Fig. 3.** Post-hoc factual, semi-factual, and counterfactual explanations for MNIST showing: (a) a *factual explanation* for a misclassification of "6" as "1", that uses a nearest-neighbor in the latent-space classed as "1", (b) a *semi-factual explanation* for the correct classification of a "9", that shows a synthetic instance with meaningful feature changes that would *not* alter its classification, and (c) a *counterfactual explanation* for the misclassification of an "8" as a "3", that shows a synthetic test-instance with meaningful feature changes that *would have been* classified as an "8" (adapted from [Kenny and Keane, 2021a]).

*weights*, *feature-importances, or predictive outcomes* – that "reflect" what the ANN has learned, is mapped to the *k*-NN retrieval and/or adaptation steps of the CBR.

- *Bipartite Division of Labor*. There is a bipartite division of labor between the ANN and CBR modules, where the former delivers prediction-accuracy, and the latter provides interpretability by explaining the ANN's outputs (in classification or regression), using factual, counterfactual, or semi-factual instances

This upgraded definition of twinning covers the range of different methods that have been developed in recent work. The substantive change to this definition lies mainly in its inclusion of new explanation strategies, moving beyond CBR's traditional, example-based explanation approach, to counterfactual and semi-factual explanations (which were not heavily researched in CBR). Notably, the definitional change expands the purview for what is mapped from the ANN to the CBR, to include information about feature importance (e.g., discriminability, exceptionality) and predictive outcomes (e.g., decision boundaries). In the next sub-section, we review the structure of the paper, before describing several specific techniques using this twinning idea (see Section 2-4).

### 1.3 Outline of Paper

In the remainder of this paper, we explore how the proposed (updated and generalized) twin-systems framework has been implemented in our work on factual, counterfactual, and semi-factual explanations for DL models (see Table 1 for a summary and Section 2-4). One of the interesting observations that arises from this review of techniques is that different explanation strategies can use different functional aspects of the DL model to achieve their aims (e.g., feature weights versus exceptional-feature information). Finally, we also discuss how a CBR counterfactual method can be used for Data Augmentation (see Section 5). As such, this paper reports three main novelties, in meeting its aims, namely:

- A new, more-general definition of twin systems
- A review of major examples of twin-systems solutions using CBR in DL for XAI and Data Augmentation
- A discussion of the basis for relating CBR and DL systems productively, for future work

In the next section, we turn to the first of the explanation strategies reviewed, using twinning for factual explanations.

**Table 1**: Summary of the types of twinning strategies adopted in recent work.

| *Explanation Strategy* | *Paper* | *Data Types* | *Functionality Mapping* |
|---|---|---|---|
| *Factual* | Kenny & Keane [2019] | tabular, image | important features |
| *Counterfactual* | Keane & Smyth [2020] | tabular | predictive outcomes |
| | Smyth & Keane [2021] | tabular | predictive outcomes |
| | Delaney *et al.* [2021] | time-series | discrimintive features |
| | Kenny & Keane [2021] | tabular, image | exceptional features |
| *Semi-factual* | Kenny & Keane [2021a] | image | exceptional features |

## 2  Twinning for Factual Explanations

Twin-systems using factual explanations were introduced by Kenny and Keane [2019] as a general method to explain neural networks locally using CBR. The framework proposes that an ANN may be abstracted in its entirety into a single proxy CBR system that mimics the ANN's predictive logic. This solution has a notable advantage over other explanation methods as CBR is non-linear (e.g., as opposed to say LIME [Ribeiro *et al.* 2016]) and can thus more accurately abstract the non-linear ANN function using only a single proxy model. Figure 3a shows an example of a factual explanation for image data – using the MNIST dataset – in which the DL model makes an incorrect classification (of a 6 as a 1), with the factual case showing the nearest neighbor (image of a 1) that explains why the DL misclassified. Kenny *et al.* [2021] reported user studies testing such misclassifications for a CNN using the actual factual examples found by the method.

Twin-system techniques differ from other factual explanation-by-example approaches in their use of feature weights. Most methods for post-hoc explanation-by-example use feature activations to locate similar training examples to a test instance (aka neuron activations in the ANN [Papernot and MacDaniel, 2018; Jeyakumar *et al.*, 2020]). In contrast, the twin-systems solution uses *feature contributions*, which weight these neuron activations by their connection weights to the predicted class. This approach has the effect of finding nearest neighbors that (i) are predicted to be in the same class as the test case, and (ii) have similarly-important features used in the prediction. Notably, this twinning operation also allows the CBR's predictive function to better mimic the ANN's function, lending credible evidence to the proposal that it has abstracted the ANN's decision-making process.

### 2.2 Factuals: Commentary

This contributions-based feature-weighting method has been widely tested on classification and regression problems involving a variety of MLP/DL models and has consistently been shown to deliver good factual explanations [Kenny & Keane, 2019, 2021b]. Indeed, Kenny & Keane [2019] showed that this feature-weighting method also out-performed the historical methods from the CBR literature (going back to the 1990s). As such, the takeaway from considering this explanation type is that the feature-contributions functionality of the DL system is a critical input to the success of the approach, though as we shall see the same may not hold for other explanation strategies.

## 3  Twinning for Counterfactual Explanations

In contrast to twin systems for factual explanations, the type of ANN functionality that has been used to compute counterfactual explanations shows significant variation. We have explored three methods that all use different aspects of the ANNs functionality, including (i) a method for time-series data using disriminative features, (ii) a pure-CBR method only needs feedback from the ANN's predictions (for decision boundaries) and (iii) a method that uses exceptional features from analyses of the ANN's data distributions. We describe the first two methods here, leaving the last to the next section, as the latter also finds semi-factuals (section 4).

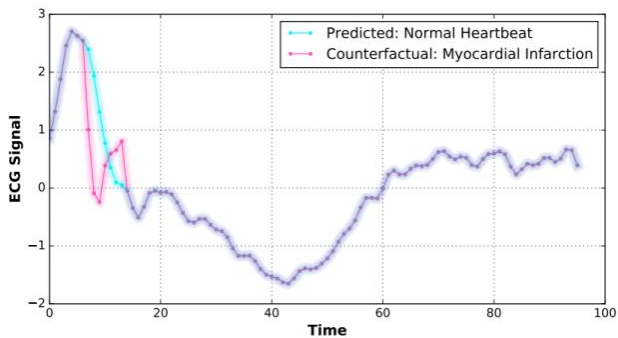

**Fig. 4:** A counterfactual instance that explains the classification of an ECG signal from [adapted from Delaney *et al*., 2021]. Here, a black-box's classification of a normal heartbeat (purple and blue line) is explained with a counterfactual showing an abnormal, heart-attack signal (purple and red line).

### 3.1 Counterfactuals Using Feature Importance

In a manner that is reminiscent of a feature contributions method used for factual explanations [Sani *et al*., 2017], counterfactual explanations for time-series datasets have used discriminative features [Delaney *et al.,* 2021]. The *Native Guide* method is a model-agnostic, case-based technique that produces plausible and diverse counterfactual explanations for black-box DL time-series classification systems [Delaney *et al.*, 2021]. Initially, *Native Guide* retrieves counterfactual solutions that already exist in the case-base (e.g., the test-case's nearest unlike neighbor or NUN, see also [Nugent *et al.*, 2009]). Next, to enhance the generation of sparse and proximate solutions, discriminative areas of the time series are located from the Deep Learner using Class Activation Mapping [Zhou *et al*., 2016]; note, if there is no access to the model's internals, then model agnostic techniques can be used (e.g., such as SHAP). These existing counterfactual cases and the discriminative-feature information are then used to guide an adaptation step that generates explanatory counterfactuals to find a discriminative, semantically-meaningful contiguous subsequence in the time series data (see Figure 4).

Results from comparative experiments on several diverse time-series datasets from the UCR archive, have indicated that *Native Guide* produces plausible and diverse explanations for a state-of-the-art, black-box CNN architecture. These studies also show that benchmark constraint-based optimization techniques [Wachter *et al*., 2018; Mothilal *et al.,* 2020] frequently failed to produce plausible counterfactual explanations in these time-series domains. We see the plausibility of the counterfactual explanations produced by *Native Guide* as being due to its generation process being grounded in existing counterfactual solutions from the training data.

## 3.2 Counterfactuals from Predictive Outcomes

Though the above use of discriminative features has been shown to be important for identifying good counterfactuals in time-series data, for tabular data another pure-CBR method works well, by using cases in the case-base that are close to the decision boundary. This instance-based counterfactual method does not require access to the internals of the DL model but rather works off feedback from the ANN's predictive outcomes (reflecting its decision boundary).

Keane and Smyth's [2020] *case-based counterfactual method* exploits known counterfactual relationships in the dataset. This instance-guided method finds a test case's nearest neighbor that takes part in a so-called *explanation case (xc)*. An explanation case is a pair of mutually-counterfactual cases that occur in the case-base that differ by at most two features (so-called "native counterfactuals"). The test case and the counterfactual case from this nearest *xc* are used to guide the generation of a new "good" counterfactual for the test case, by adapting the test-case's features with the (at most) two difference-features from the *xc's* counterfactual cases. By construction, these new counterfactuals are guaranteed to be contain feature values that naturally occur in the problem space, rather than interpolations of feature values that may implausible (such as those that can be produced by Wachter *et al.'s* [2018] method).

In a series of tests conducted over a wide range of datasets, this CBR method has been shown to generate plausible counterfactuals, often involving counterfactual relationships that are closer than those found in native counterfacutals in the dataset [Keane and Smyth, 2020; Smyth and Keane, 2021]. Interestingly, recent user testing has shown that people prefer counterfactual explanations with 2-3 feature differences, supporting a key assumption of this method [Förster *et al.*, 2020]. This technique has also been used for Data Augmentation, where again good results are found for 2-difference counterfactuals (see section 5).

## 3.3 Counterfactuals: Issues & Comments

These CBR solutions for counterfactual explanations of DL models, illustrate how twinning can change depending on the explanation strategy being computed. For factual explanations, the twinning process relies on an analysis of feature weights, that are then re-used in the CBR retrieval step. For counterfactual explanations, an analysis of discriminative-features can benefit the generation of plausible counterfactuals in the time-series domain. However, we also saw that counterfactual explanations can also be generated with *just* a knowledge of the DL model's predictions. However, it should be noted, that this sort of approach will always be approximate; the counterfactual technique is arguably just "getting by" with minimal information from the twinned DL. Interestingly, much of the discussion in the counterfactual literature hinges on this question; namely, how well a technique can do without more knowledge of the twinned-model and/or the domain (e.g., distributional information, causal knowledge of the domain, feature-importance information; see [Karimi *et al.*, 2020] for discussions). Indeed, in the next section, when we review our final method for semi-factuals (and counterfactuals), we will see another type of information – about feature exceptionality – can also be used to compute counterfactuals.

## 4 Twinning for Semi-Factuals

We have seen that CBR solutions for factual and counterfactual explanations in twin systems can be guided by some analysis of the features in the DL model. There is a further option that specifically hinges on using *exceptional features* that can generate semi-factuals (and counterfactuals). This is best understood by first considering the relationship between semi-factuals and counterfactuals.

Recall, that the counterfactual explains by telling the user what will change the outcome (e.g., yield will be higher if you use more fertilizer); a good counterfactual will take the user just over the decision boundary, into the closest possible world in which the outcome is different [Lewis, 1973]. The semi-factual explains by telling the user about how feature-values can change *without* producing a different outcome (e.g., yield will be the same even if you double fertilizer use); a good semi-factual takes the user to a point "just before" the decision boundary, it is akin to telling the user about the headroom in a feature *before* a counterfactual change occurs. Hence, semi-factuals (which have received little attention in the literature) tend to be computed relative to counterfactuals; the semi-factual can be viewed as a perturbation of the counterfactual that leaves the class *as is*.

### 4.1 Semi-factuals from Exceptional Features

This is the approach adopted by Kenny and Keane's [2021a] Plausible Exceptionality-Based Contrastive Explanations (PIECE) method from which they generate semi-factuals and counterfactuals based on an analysis the feature-distributions of classes found in the DL model; specifically, they find exceptional features in the counterfactual class and then perturb the test case using these features. When the exceptional features used in this perturbation are rank ordered based on exceptionality and applied successively, a semi-factual is generated "on the way" to finding the counterfactual; the best semi-factual is the instance generated from perturbing exceptional features, just before crossing the decision boundary to generate the counterfactual.

PIECE works by identifying "exceptional features" in a test instance with reference to the training distribution; that is, features of a low probability in the counterfactual class are modified to be values that occur with a high probability in that class. For example, when a CNN has been trained on the MNIST dataset and a test image labelled as "8" is misclassified as "3", the exceptional features (i.e., low probability features in the counterfactual class 8) are identified in the extracted feature layer of the CNN via statistical modelling (i.e., a *hurdle model* to model ReLU activations) and modified to be their expected statistical values for the 8-counterfactual-class (see Figures 3b and 3c).

Depending on the number of exceptional features changed, PIECE will produce a semi-factual or counterfactual. Note, at present, PIECE uses a generative model (i.e., a GAN) to help produce these synthetic cases (as it helps guarantee plausibility), but an earlier version of the system simply used a *k*-NN to find the closest instance in the training data to the modified test instance.

### 4.2 Semi-factuals: Issues & Comments

So, here we can see that semi-factuals (and counterfactuals) can be found for DL-CBR twins relying on a somewhat different feature-analysis (based on exceptionality). The core difference with PIECE over the earlier methods discussed is that the case base is not used directly to retrieve a case for explanation, but rather its distributional properties are summarized to inform synthetic explanation generation, though a nearest case could just as easily be used.

## 5 Data Augmentation Using Counterfactuals

Thus far, we have seen that several productive links can be made between Deep Learning and CBR in the XAI field. But, Leake & Crandall [2020] sketched a broader canvas, arguing that there should be many points of contact between DL and CBR (e.g., with respect to data-generation problems). Recently, we have considered one such avenue in Data Augmentation. It is well-known that DL models typically require large datasets to be successful and several Data Augmentation methods have emerged to solve this problem [Antoniou *et al.,* 2017; Chawla *et al.,* 2002; Shin *et al.,* 2018]. In recent work, we have found that the case-based counterfactual method [Keane and Smyth, 2020] generates useful synthetic cases to augment datasets, to help them deal with the problems caused by the "concept drift" associated with climate change.

Temraz *et al.* [2021] examined the prediction problems faced by a grass-growth prediction model for precision agriculture dealing with climate-disruptive events [Kenny *et al.*, 2019, 2020]; when climate disruption occurs – as in the very hot weather in Europe in the summer of 2018 – past cases become less useful and predictive accuracy drops. Temraz *et al.* [2021] took an historical case-base of grass-growth records from 6,000+ Irish dairy farms, for the years 2013-2016, and generated counterfactuals along a climate-defined decision boundary between "normal climate" and "outlier climate" cases. They generated >2,500 synthetic outlier cases from the 2013-2016 dataset using the case-based counterfactual method and then checked the predictive accuracy of the model using these synthetic cases, on the climate-disrupted year of 2018. They found significant improvements in prediction accuracy, when the dataset was augmented with these outlier-counterfactuals; specifically in those months in which climate-disruption occurred. Temraz *et al.* [2021] argue that the counterfactual method does something akin to creating a local adaptation-rule that is then used to generate new, synthetic cases. Notably, they found that the results improved most for counterfactuals generated by the case-based counterfactual method; that is, a benchmark, optimization-based counterfactual method does not show similar improvements in accuracy [Mothilal *et al.,* 2020]. Furthermore, Temraz and Keane [2021] have shown that these counterfactual benefits for data augmentation generalize to a wide range of datasets, for many different classifiers (including an MLP); notably, the counterfactual method also did better than the popular SMOTE data augmentation technique [Chawla *et al.,* 2002]. In short, this initial work suggests that the CBR benefits shown for DL in XAI, extend to Data Augmentation too.

## 6 Conclusions, Issues, and Futures

In this paper, we have explored a line of work showing interesting synergies between DL and CBR focused on the problem of XAI (and Data Augmentation). All of these solutions conform to the revised definition of twin systems outlined earlier.

For us, the most notable aspect of this work is the extent to which the case-based approach delivers good solutions in these problem domains. Indeed, in many cases, the case-based solution appears to be among the best solutions in the field. Furthermore, different CBR techniques can operate off different functional aspects of DL models (i.e., feature contributions, discriminative features, exceptionality, predictive outcomes). Indeed, in some cases – as in counterfactuals – there seems to be some flexibility with respect to what functionalities need to be mapped from DL to CBR, raising the issue of which are optimal for the best outcomes. A final takeaway, perhaps, is the richness of the synergies that potentially exist in this DL-CBR interaction; here, we have looked in depth at one area and just touched on another. When one considers other potential problem areas, a bright and interesting future seems to await researchers in exploring this DeepCBR concept.

### Acknowledgements

This publication has emanated from research conducted with the financial support of (i) Science Foundation Ireland (SFI) to the *Insight Centre for Data Analytics* under Grant Number 12/RC/2289_P2, and (ii) SFI and the Department of Agriculture, Food and Marine on behalf of the Government of Ireland to the *VistaMilk SFI Research Centre* under Grant Number 16/RC/3835. For open access, the author has applied a CC BY public copyright license to any author accepted manuscript version arising from this submission.

### References

[Antoniou *et al.,* 2017] Antoniou, Antreas, Amos Storkey, and Harrison Edwards. Data augmentation generative adversarial networks. *arXiv:1711.04340*. 2017.

[Byrne, 2019] Ruth Byrne. Counterfactuals in explainable artificial intelligence (XAI): Evidence from human reasoning. In *IJCAI-19*, pages 6276–6282, 2019.

[Chawla *et al.,* 2002] Nitesh Chawla, Kevin Bowyer, Lawrence Hall, and W Philip Kegelmeyer. SMOTE:

Synthetic minority over-sampling technique. *Journal of Artificial Intelligence Research*, 16, 321-357, 2002.

[Delaney *et al.*, 2021] Eoin Delaney, Derek Greene and Mark T. Keane. Instance-based counterfactual explanations for time series classification. In *ICCBR-21,* Springer, 2021

[Förster *et al.,* 2020] Maximilian Förster, Mathias Klier, Kilian Kluge, and Irina Sigler. Fostering human agency. In *ICIS-2020,* paper 1963, 2020.

[Karimi *et al.*, 2020] Amir-Hossein Karimi, Gilles Barthe, B Schölkopf and I Valera. A survey of algorithmic recourse. *arXiv preprint:2010.04050*, 2020.

[Keane and Kenny, 2019] Mark T. Keane and Eoin M. Kenny. How case-based reasoning explains neural networks. In *ICCBR-19,* pp. 155-171, Springer, 2019.

[Keane et al., 2021] Mark T. Keane, Eoin M. Kenny, Eoin Delaney and Barry Smyth. If only we had better counterfactual explanations. In *IJCAI-21*, 2021.

[Keane and Smyth, 2020] Mark T. Keane and Barry Smyth. Good counterfactuals and where to find them. In *ICCBR-20*, pages 163-178. Springer, 2020.

[Kenny and Keane, 2019] Eoin M. Kenny and Mark T. Keane. Twin-systems for explaining ANNs using CBR. In *IJCAI-19*, pages 2708-2715, 2019.

[Kenny and Keane, 2021a] Eoin M. Kenny and M T Keane. On generating plausible counterfactual & semi-factual explanations for deep learning. In *AAAI-21*, pages 11575-11585, 2021.

[Kenny and Keane, 2021b] Eoin M. Kenny and Mark T. Keane. Explaining deep learning with examples 2021.

[Kenny *et al.*, 2019] Eoin M. Kenny, Elodie Ruelle, Anne Geoghegan, Laurence Shalloo, Micheál O'Leary, Michael O'Donovan, and Mark T. Keane. Predicting grass growth for sustainable dairy farming. In *ICCBR-19,* pp. 172-187. Springer, Cham, 2019.

[Kenny e*t al.*, 2021] Eoin M. Kenny, Courtney Ford, Molly Quinn and Mark T. Keane. Post hoc explanations for deep learning. *Artificial Intelligence*, 294, 2021.

[Leake and Crandall, 2020] David Leake and David Crandall. On bringing Case-based Reasoning methodology to Deep Learning. In *ICCBR-20*, pages 343-348. Springer, Cham, 2020.

[Leake and McSherry, 2005] David Leake and David McSherry. Introduction to the special issue on explanation in case-based reasoning. *Artificial Intelligence Review, 24(2)*, 103-108 2005.

[Lewis, 1973/2013] David Lewis. *Counterfactuals*. John Wiley & Sons, London, 1973/2013.

[McCloy and Byrne, 2002] Rachel McCloy and Ruth M.J. Byrne. Semifactual "even if" thinking. *Thinking & Reasoning* 8, pages 41-67, 2002.

[Lipton, 2018] Zachary C Lipton. The mythos of model interpretability. *Queue* 16(3), pages 31-57, 2018.

[Miller, 2019] Tim Miller. Explanation in Artificial Intelligence: Insights from the social sciences. *Artificial intelligence*, 267, pages 1-38, 2019.

[Mothilal *et al*., 2020] Ramaravind K Mothilal, Amit Sharma, and Chenhao Tan. Explaining machine learning classifiers through diverse counterfactual explanations. In *FAT*2020*, pages 607– 617, 2020.

[Nugent *et al.*, 2009] Conor Nugent, Dónal Doyle, and Pádraig Cunningham. Gaining insight through case-based explanation. *Journal of Intelligent Information Systems*, 32, pages 267-295, 2009.

[Papernot and McDaniel, 2018] , Nicolas Papernot and Patrick McDaniel. Deep k-nearest neighbors: Towards confident, interpretable and robust deep learning. *arXiv preprint arXiv:1803.04765*, 2018.

[Ribeiro *et al*., 2016] Marco Tulio Ribeiro, Sameer Singh and Carlos Guestrin. Why should I trust you?: Explaining the predictions of any classifier. In *SIGKDD-16*, pages1135-1144, 2016.

[Sani *et al*., 2017] Sadiq Sani, Nirmalie Wiratunga, and Stewart Massie. Learning deep features for kNN-based human activity recognition. In *ICCBR-17 Workshop Proceedings*. Springer, 2017.

[Shin *et al.,* 2018] Hoo-Chang Shin, Neil Tenenholtz, Jameson Rogers, Christopher Schwarz, Matthew Senjem, Jeffrey Gunter, Katherine Andriole, and Mark Michalski. Medical image synthesis for data augmentation and anonymization using generative adversarial networks. In *International workshop on simulation and synthesis in medical imaging*, pages 1-11. Springer, Cham, 2018.

[Shin *et al*., 2000] C-K. Shin, Yun, U., Kim,, H.K., and Park, S.C. A hybrid approach of neural network and memory-based learning to data mining, *IEEE Trans. on Neural Networks*, 11(3), pages 637–646, 2000.

[Sørmo *et al.*, 2005] F. Sørmo, Jorg Cassens, and Aamodt, A. Explanation in case-based reasoning. *Artificial Intelligence Review, 24(2),* pages 109-143, 2005.

[Smyth and Keane, 2021] Barry Smyth and Mark T. Keane. A Few Good Counterfactuals. *arXiv Preprint :2101.09056*, February 2021.

[Temraz and Keane, 2021] Mohammed Temraz and Mark T. Keane. Using counterfactuals for handling the class imbalance problem, In submission, 2021.

[Temraz *et al.,* 2021] Mohammed Temraz, Eoin Kenny, Elodie Ruelle, Laurence Shalloo, Barry Smyth, and Mark T. Keane. Handling climate change using counterfactuals. *ICCBR-21,* Springer, 2021

[Wachter *et al.*, 2018] Sandra Wachter, Brent Mittelstadt, and Chris Russell. Counterfactual explanations without opening the black box: Automated decisions and the GDPR. *Harv. JL & Tech., 31*, pp. 841, 2018.

[Zhou *et al*., 2016] Bolei Zhou, Aditya Khosla, Agata Lapedriza, Aude Oliva, and Antonio Torralba. Learning deep features for discriminative localization. In *IEEE Conference on Computer Vision and Pattern Recognition*, pages 2921–2929, 2016.